# Applying Winnow to Context-Sensitive Spelling Correction

**Andrew R. Golding**
Mitsubishi Electric Research Labs
201 Broadway
Cambridge MA 02139
USA
golding@merl.com

**Dan Roth**
Dept. of Appl. Math. & CS
Weizmann Institute of Science
Rehovot 76100
Israel
danr@wisdom.weizmann.ac.il

## Abstract

Multiplicative weight-updating algorithms such as Winnow have been studied extensively in the COLT literature, but only recently have people started to use them in applications. In this paper, we apply a Winnow-based algorithm to a task in natural language: context-sensitive spelling correction. This is the task of fixing spelling errors that happen to result in valid words, such as substituting *to* for *too*, *casual* for *causal*, and so on. Previous approaches to this problem have been statistics-based; we compare Winnow to one of the more successful such approaches, which uses Bayesian classifiers. We find that: (1) When the standard (heavily-pruned) set of features is used to describe problem instances, Winnow performs comparably to the Bayesian method; (2) When the full (unpruned) set of features is used, Winnow is able to exploit the new features and convincingly outperform Bayes; and (3) When a test set is encountered that is dissimilar to the training set, Winnow is better than Bayes at adapting to the unfamiliar test set, using a strategy we will present for combining learning on the training set with unsupervised learning on the (noisy) test set.

## 1 INTRODUCTION

Multiplicative weight-updating algorithms such as Winnow [Littlestone, 1988] and Weighted Majority [Littlestone and Warmuth, 1994] have been studied extensively in the COLT literature. Theoretical analysis has shown that they have exceptionally good behavior in the presence of irrelevant attributes, noise, and even a target function changing in time [Littlestone, 1988; Littlestone and Warmuth, 1994; Herbster and Warmuth, 1995]. We address these claims empirically by applying a new algorithm which combines variants of Winnow and Weighted Majority to a large-scale real-world task: context-sensitive spelling correction.

Context-sensitive spelling correction is the task of fixing spelling errors that result in valid words, such as *It's not to late*, where *too* was mistakenly typed as *to*. These errors account for anywhere from 25% to over 50% of observed spelling errors [Kukich, 1991]; yet they go undetected by conventional spell checkers, such as Unix *spell*, which only flag words that are not found in a word list. The challenge of this task (and related natural language tasks) for Machine Learning is to characterize the contexts in which a word can (or cannot) occur in terms of features. The problem is that there is a multitude of features one might use: features that test for the presence of a particular word nearby the target word; features that test the pattern of parts of speech around the target word; and so on. For the tasks we will consider, the number of features ranges from a few hundred to over 10,000.[1] Most machine-learning algorithms, such as C4.5, do not scale up well to such large feature sets; the task therefore provides an excellent testbed for studying the performance of multiplicative weight-updating algorithms on a real-world task with large numbers of features.

To evaluate the proposed Winnow-based algorithm, which we call *WinnowS*, we compare it against *Bayes* [Golding, 1995], a statistics-based method that is among the most successful methods tried for the problem. We first compare WinnowS and Bayes using the heavily-pruned feature set that Bayes normally uses (typically 10–1000 features). WinnowS is found to perform comparably to Bayes under this condition. When the full, unpruned feature set is used, however, WinnowS comes into its own, achieving substantially higher accuracy than it achieved in the pruned condition, and better accuracy than Bayes achieved in either condition.

---

[1] We have tested successfully with up to 40,000 features, but the results reported here use up to 11,000.

We then address the issue of dealing with a test set that is dissimilar to the training set. This arises in context-sensitive spelling correction because patterns of word usage can vary widely across documents; thus the test and training documents may be quite different. To deal with this, we adopt a strategy of combining learning on the training set with unsupervised learning on the (noisy) test set. Using this strategy, WinnowS is found to be convincingly better than Bayes at adapting to unfamiliar test sets.

The rest of the paper is organized as follows: the next sections describe the task of context-sensitive spelling correction, and the Bayesian method that has been used for it. The Winnow-based approach to the problem is then introduced. The experiments on WinnowS and Bayes are presented. The final section concludes.

## 2 CONTEXT-SENSITIVE SPELLING CORRECTION

Context-sensitive spelling correction is taken here to be a task of word disambiguation. The ambiguity among words is modelled by *confusion sets*. A confusion set $C = \{W_1, \ldots, W_n\}$ means that each word $W_i$ in the set is ambiguous with each other word. Thus if $C = \{hear, here\}$, then when we see an occurrence of either *hear* or *here* in the target document, we take it to be ambiguous between *hear* and *here*; the task is to decide from the context which one was actually intended. Acquiring confusion sets is an interesting problem in its own right; in the work reported here, however, we will just take our confusion sets from the list of "Words Commonly Confused" in the back of the Random House dictionary [Flexner, 1983].

The Bayesian and Winnow-based methods for spelling correction will be described below in terms of their operation on a single confusion set; that is, we will say how they disambiguate occurrences of words $W_1$ through $W_n$. The methods handle multiple confusion sets by applying the same technique to each confusion set independently.

## 3 BAYESIAN APPROACH

A number of methods have previously been tried for context-sensitive spelling correction, including trigrams [Mays et al., 1991], Bayesian classifiers [Gale et al., 1993], decision lists [Yarowsky, 1994], Bayesian hybrids [Golding, 1995], and, more recently, a combination of trigrams and Bayesian hybrids [Golding and Schabes, 1996]. The Bayesian hybrid method, which we call Bayes, has been among the most successful, and is thus the method we adopt here as the benchmark for comparison with WinnowS. Bayes has been described elsewhere [Golding, 1995], and so will only be briefly reviewed here; however, the version used here uses an improved smoothing technique, which is mentioned briefly below.

To disambiguate words $W_1$ through $W_n$, Bayes starts by learning features that characterize the context in which each $W_i$ tends to occur. It uses two types of features: *context words* and *collocations*. Context-word features test for the presence of a particular word within $\pm k$ words of the target word; collocations test for a pattern of up to $\ell$ contiguous words and/or part-of-speech tags[2] around the target word. In the experiments reported here, $k$ was set to 10 and $\ell$ to 2. Examples of features for the confusion set $\{weather, whether\}$ include:

(1)  *cloudy* within $\pm 10$ words
(2)  __ to VERB

where (1) is a context-word feature that tends to imply *weather*, and (2) is a collocation that checks for the pattern "to VERB" immediately after the target word, and tends to imply *whether* (as in *I don't know whether to laugh or cry*).

Bayes learns these features from a training corpus of correct text. Each time a word in the confusion set occurs, Bayes proposes every feature that matches that context (one context-word feature for every distinct word within $\pm k$ words, and one collocation for every way of expressing a pattern of up to $\ell$ contiguous elements). After working through the whole training corpus, Bayes tallies the number of times each feature was proposed. It then prunes features for two reasons: (1) the feature occurred in practically none or all of the training instances (specifically, it had fewer than 10 occurrences or fewer than 10 non-occurrences); or (2) the presence of the feature is not significantly correlated with the identity of the target word (determined by a chi-square test at the 0.05 significance level).

The set of learned features is used at run time to classify an occurrence of a word in the confusion set. All the features are compared against the target occurrence; let $\mathcal{F}$ be the set of features that match. Suppose for a moment that we were applying a *naive* Bayesian approach. We would then calculate the probability that each word $W_i$ in the confusion set is the correct identity of the target word, given that we have observed features $\mathcal{F}$, by using Bayes' rule with the independence assumption:

$$P(W_i|\mathcal{F}) = \left(\prod_{f \in \mathcal{F}} P(f|W_i)\right) \frac{P(W_i)}{P(\mathcal{F})}$$

---

[2]Each word in the sentence is tagged with its *set* of possible part-of-speech tags, obtained from a dictionary. For a tag to match a word, the tag must be a member of the word's tag set.

where each probability on the right-hand side is calculated by a maximum-likelihood estimate (MLE) over the training set. We would then pick as our answer the $W_i$ with the highest $P(W_i|\mathcal{F})$. The method presented here differs from the naive approach in two respects: first, it does not assume independence among features, but rather has heuristics for detecting strong dependencies, and resolving them by deleting features until it is left with a reduced set $\mathcal{F}'$ of (relatively) independent features, which are then used in place of $\mathcal{F}$ in the formula above. Second, to estimate the $P(f|W_i)$ terms, rather than using a simple MLE, it performs *smoothing* by interpolating between the MLE of $P(f|W_i)$ and the MLE of the unigram probability, $P(f)$. These enhancements greatly improve the performance of Bayes over the naive Bayesian approach.

## 4  WINNOW-BASED APPROACH

The approach presented in this section is being developed as part of a research program in which we are trying to understand how networks of simple and slow neuron-like elements can encode a large body of knowledge and perform a wide range of interesting inferences almost instantaneously [Khardon and Roth, 1994; Roth, 1996a]. In particular, we investigate this question in a system developed for the purpose of learning knowledge representations for natural language understanding tasks [Roth, 1996b]. In the following we briefly present the general approach and then concentrate on the task at hand, context-sensitive spelling correction.

The approach developed is influenced by the Neuroidal system suggested in [Valiant, 1994]. The system consists of a very large number of items, in the range of $10^5$. These correspond to so-called high-level concepts, for which humans have words, as well as to lower-level predicates from which higher-level ones are composed. The knowledge representation is a large network of threshold gates, in which every concept is defined as a function of other nodes. More specifically, every concept is represented as a *cloud* of nodes in the network. While each node in this cloud learns its dependence on other nodes in the network autonomously, all the cloud's members take part in any decision with respect to the concept. A continuous learning approach [Valiant, 1995] is used to learn and maintain[3] the knowledge representation. Each interaction with the world, e.g., reading a sentence of text, is viewed as a positive example of a few of these items and a negative example for all the others. Each example is thus used once by all the items to learn and refine their definition in terms of the others [Valiant, 1995; Roth, 1995], and is then discarded. Local learning algorithms are used at each node: a variant of Littlestone's Winnow algorithm [Littlestone, 1988] is used by each node to learn its dependence on other nodes, but different members of a concept cloud run this algorithm with different parameters. A decision with respect to the concept represented by the cloud is reached via a variant of the Weighted Majority algorithm [Littlestone and Warmuth, 1994].

---

[3]For the purpose of this experimental study, except when specifically mentioned, we do not update the knowledge representation while testing, even though the information is available to the learning algorithm. This is done to provide a fair comparison with the Bayesian method which is a batch approach.

For the current task, we keep a node for every word which appears in a sentence read by the system, as well as every collocation feature. Edges are added only between nodes that happen to be active in the same sentence. Given an example for the confusion set {*desert, dessert*}, all the nodes that correspond to active features in the example are activated. Then an update stage occurs, in which each member of the *desert* and *dessert* clouds updates its representation as a function of the active nodes. For the sake of this update, if the sentence consists of the word *desert* then the example is viewed as a positive example for the *desert* nodes and a negative example for the *dessert* nodes. When testing, essentially the same process is done, except that the prediction is done at the concept (cloud) level. We predict *desert* if the total weighted activation of the members of this cloud outweighs that of *dessert*.

### 4.1  Using Winnow

At each node we use a version of Littlestone's Winnow2 algorithm [Littlestone, 1988]. If the number of features (variables) of the target function is $n$, the algorithm keeps an $n$-dimensional weight vector of positive weights. The algorithm has three parameters: a threshold $\theta$, and two update parameters, a *promotion* parameter $\alpha > 1$ and a *demotion* parameter $0 < \beta < 1$. For a given instance $(x_1, \ldots, x_n)$ the algorithm predicts 1 iff $\sum_{i=1}^{n} w_i x_i > \theta$, where $w_i$ is the weight on the edge connecting $x_i$ to the target node. Thus the hypothesis of this algorithm is a linear threshold function over $\{0,1\}^n$. The algorithm updates its hypothesis only when a mistake is made. If the algorithm predicts 0 and the received label is 1 (positive example) then for all indices $i$ such that $x_i = 1$, the weight $w_i$ is replaced by a larger weight $\alpha \cdot w_i$. If the algorithm predicts 1 and the received label is 0 (negative example) then for all indices $i$ such that $x_i = 1$, the weight $w_i$ is replaced by a smaller weight $\beta \cdot w_i$. In both cases, if $x_i = 0$, its weight remains unchanged.

Instead of representing each example as a bit vector, we represent it simply as a list of its *active* attributes (all the 1 bits), leaving unmentioned all the attributes the example does not have (as in [Blum, 1992]).

Consider for example a confusion set $\{W_1, W_2\}$. Given a sentence, it is first translated into a list of *active* features, as in Section 3. Assume $W_1$ is in this list. Then the example is used as a *positive* example for each $W_1$ node and a *negative* example for the $W_2$ nodes. Let $\{w_i\}_{i=1}^{k}$ be the set of weights on incoming edges of a node in $W_1$. Given the list of active attributes, the node predicts 1 iff $\sum_{i=1}^{k} w_i x_i > \theta$, where $x_i = 1$ when $x_i$ is present in the list of active attributes, and $x_i = 0$ otherwise. Then, the weights $\{w_i\}_{i=1}^{k}$ are updated based on the Winnow update rule and the corresponding value of the attributes and the target node.

A further addition we make in the Winnow variant used is that we drop poorly-performing attributes, whose weight falls too low relative to the highest incoming weight of the target node. Intuitively, this could result in an algorithm that actually speeds up as it learns. Also, we try to identify cases where conjunctions of the primitive features measured are actually the relevant features. This is done by keeping track of dynamically changing blocks of attributes that behave (almost) the same. Thus we keep track of fewer features than we actually measure.

In the experiments presented here, we use $\theta = 1$ and promotion parameter $\alpha = 1.5$. Cloud members use different demotion parameters which vary between 0.5 and 0.9. The initial weights were $1/d$, where $d$ is a typical number of active features in an example.

### Properties

Winnow was shown to learn efficiently any linear threshold function [Littlestone, 1988]. These are functions $f : \{0,1\}^n \to \{0,1\}$ for which there exist real weights $w_1, \ldots, w_n$ and a real threshold $\theta$ such that $f(x_1, \ldots, x_n) = 1$ iff $\sum_{i=1}^{n} w_i x_i \geq \theta$. In particular, these functions include Boolean disjunctions and conjunctions on $k \leq n$ variables and $r$-of-$k$ threshold functions ($1 \leq r \leq k \leq n$). The key feature of Winnow is that its mistake bound grows linearly with the number of *relevant* attributes and only logarithmically with the total number of attributes $n$. Using the version mentioned above, in which we do not keep all the variables from the beginning, but add variables when necessary, the number of mistakes made on disjunctions and conjunctions is linear with the size of the largest example seen and with the number of relevant attributes, and is independent of the total number of attributes in the domain [Blum, 1992].

Winnow was analyzed in the presence of various kinds of noise, and in cases where no linear-threshold function can make perfect classifications [Littlestone, 1991]. It is proven, under some assumptions on the type of noise, that Winnow still learns correctly, while retaining its abovementioned dependence on the number of total and relevant attributes. (See [Kivinen and Warmuth, 1995] for a thorough analysis of multiplicative update algorithms versus additive update algorithms.) The algorithm makes no independence or any other assumptions on the attributes, in contrast to Bayesian predictors which are commonly used in statistical NLP. This condition was recently investigated experimentally (but on simulated data) [Littlestone, 1995]. It was shown that redundant attributes dramatically affect a Bayesian predictor, while superfluous independent attributes are not as dramatic, and have an effect only when the number of attributes is very large (on the order of 10,000).

Winnow is a mistake-driven algorithm; that is, it updates its hypothesis only when a mistake is made. Intuitively, this makes the algorithm more sensitive to the relationships among the attributes — relationships that may go unnoticed by an algorithm that is based on counts accumulated separately for each attribute. This is crucial in the analysis of the algorithm and has been shown to be crucial empirically as well [Littlestone, 1995].

### 4.2 Weighted Majority

Every concept (a member of a confusion set) is represented as a *cloud* of nodes in the network. While each node in this cloud is running Winnow and learns its dependence on other nodes in the network autonomously, all the cloud's members take part in any decision with respect to the concept. In prediction, each member of the cloud can be viewed as an *expert*. The global algorithm uses the expert's activation level to make a prediction.

Different members of the cloud run the Winnow variant described above, each with a distinct *demotion* parameter. The intuition is that the words in different confusion sets may overlap to various degrees, and should be penalized accordingly for a mistaken prediction. For example, in the case of $\{among, between\}$, there is a considerable overlap in the usage of the words, and therefore, a sentence in which the word *among* appears is a negative example for *between* only with some small probability. On the other hand, $\{weather, whether\}$ have disjoint usages, and every occurrence of *weather* is a negative example of *whether* with certainty. Thus, following a mistake, in the former case we want to demote the weights by a smaller ratio than in the latter. Running Winnow with various demotion parameters in parallel allows the algorithm to select by itself the best setting of parameters for each target word.

The question of combining the predictions of experts in a way that minimizes the total number of mistakes has been studied extensively [Littlestone and Warmuth, 1994; Cesa-Bianchi et al., 1995; Cesa-Bianchi et al., 1994]. The general approach is to assign to each expert a weight of the form $\gamma^m$, where $0 < \gamma < 1$ is a constant and $m$ is the total number of mistakes incurred by the expert so far. The essential property is that the weight of experts making many mistakes rapidly disappears. In our case, after testing a variety of weighting and voting methods, we decided to use a variant of the abovementioned weighted majority scheme in which we start with $\gamma = 1$ and decrease its value with the number of examples seen, to avoid weighing mistakes of the initial hypotheses too heavily.

In testing, given the list $x$ of active attributes, each node in a concept cloud evaluates $\sum_{i=1}^{k} w_{ij} x_i$, where $\{w_{ij}\}_{i=1}^{k}$ is the set of incoming weights to the node. We would then pick as our answer the concept with the highest weighted sum of activation level, summed over the concept's nodes.

## 5 EXPERIMENTAL RESULTS

To understand the performance of WinnowS on the task of context-sensitive spelling correction, we start by comparing it with Bayes using the pruned set of features that Bayes normally uses (see Section 3). This evaluates WinnowS purely as a method of combining evidence from multiple features. An important claimed strength of the Winnow-based approach, however, is the ability to handle large numbers of features. We tested this by (essentially) disabling pruning, resulting in tasks with over 10,000 features, and seeing how WinnowS and Bayes scale up.

The preceding experiments drew the training and test sets from the same population, following the traditional PAC-learning assumption. This assumption may be unrealistic for the task at hand, however, where a system may encounter a target document quite unlike those seen during training. To check whether this was in fact a problem, we tested the across-corpus performance of the methods. We found it was indeed noticeably worse than within-corpus performance.

To deal with this problem of unfamiliar test sets, we tried a strategy of combining learning on the training set with unsupervised learning on the (noisy) test set. We tested how well WinnowS and Bayes were able to perform on an unfamiliar test set using this strategy.

The sections below present these experiments, interleaved with discussion.

### 5.1 Pruned versus Unpruned

The first step of the evaluation was to test WinnowS under the same conditions that Bayes normally runs under — i.e., using Bayes' usual pruned set of features. We used a random 80-20 split (by sentence) of the 1-million-word Brown corpus [Kučera and Francis, 1967] for the training and test sets. Bayes was trained first, to learn its pruned set of features; each algorithm was then tested, using this set of features, on 21 confusion sets taken from the list of "Words Commonly Confused" in the back of the Random House dictionary [Flexner, 1983]. The results appear in the 'Pruned' columns of Table 1. Although for a few confusion sets, one algorithm or the other does better, overall WinnowS performs comparably to Bayes.

The preceding comparison shows that WinnowS is a credible method for this task; but it does not test the claimed strength of Winnow — the ability to deal with large numbers of features. To test this, we modified Bayes to do only minimal pruning of features: features were pruned only if they occurred exactly once in the training set (such features are both extremely unlikely to afford good generalizations, and extremely numerous). The hope is that by considering the full set of features, we will pick up many "minor cases" — what Holte et al. [1989] have called "small disjuncts" — that are normally filtered out by the pruning process. The results are shown in the 'Unpruned' columns of Table 1. It can be seen that WinnowS almost always improves, sometimes markedly, going from the pruned to the unpruned condition; moreover, it outperforms Bayes for every confusion set except one, where it ties. The results below will all focus on the behavior of the algorithms in the unpruned case.

### 5.2 Degradation for Unfamiliar Test Sets

The preceding experiment assumed that the training set will be representative of the test set. For context-sensitive spelling correction, however, this assumption may be overly strong; this is because word usage patterns vary widely from one author to another, or even one document to another. For instance, an algorithm may have been trained on one corpus to discriminate between *desert* and *dessert*, but when tested on an article about the Persian Gulf War, will be unable to detect the misspelling of *desert* in *Operation Dessert Storm*. To check whether this is in fact a problem, we tested the across-corpus performance of the algorithms: we again trained on 80% of Brown, but tested on a randomly-chosen 40% of the sentences of WSJ, a 3/4-million-word corpus of articles from The Wall Street Journal [Marcus et al., 1993]. The algorithms were run in the unpruned condition. The results appear in Table 2. It can be seen that both algorithms degraded on most confusion sets to varying degrees.

| Confusion set | Test cases | Pruned | | | Unpruned | | |
|---|---|---|---|---|---|---|---|
| | | Features | Bayes | WinnowS | Features | Bayes | WinnowS |
| accept, except | 50 | 78 | 88.0 | 87.8 | 849 | 92.0 | 92.0 |
| affect, effect | 49 | 36 | 98.0 | 100.0 | 842 | 98.0 | 100.0 |
| among, between | 186 | 145 | 75.3 | 75.8 | 2706 | 78.5 | 84.4 |
| amount, number | 123 | 68 | 74.8 | 73.2 | 1618 | 80.5 | 86.2 |
| begin, being | 146 | 84 | 95.2 | 89.7 | 2219 | 94.5 | 95.9 |
| cite, sight, site | 34 | 24 | 76.5 | 64.7 | 585 | 73.5 | 82.4 |
| country, county | 62 | 40 | 88.7 | 90.0 | 1213 | 91.9 | 96.8 |
| its, it's | 366 | 180 | 94.5 | 96.4 | 4679 | 95.9 | 99.2 |
| lead, led | 49 | 33 | 89.8 | 87.5 | 833 | 85.7 | 93.9 |
| fewer, less | 75 | 6 | 96.0 | 94.4 | 1613 | 92.0 | 93.3 |
| maybe, may be | 96 | 86 | 90.6 | 84.4 | 1639 | 95.8 | 97.9 |
| I, me | 1225 | 1161 | 97.8 | 98.2 | 11625 | 98.3 | 99.5 |
| passed, past | 74 | 141 | 89.2 | 90.5 | 1279 | 90.5 | 95.9 |
| peace, piece | 50 | 67 | 74.0 | 72.0 | 992 | 92.0 | 94.0 |
| principal, principle | 34 | 38 | 85.3 | 84.8 | 669 | 85.3 | 94.1 |
| quiet, quite | 66 | 41 | 95.5 | 95.4 | 1200 | 89.4 | 90.9 |
| raise, rise | 39 | 24 | 79.5 | 74.3 | 621 | 84.6 | 89.7 |
| than, then | 514 | 857 | 93.6 | 96.9 | 6813 | 93.4 | 97.9 |
| their, there, they're | 850 | 946 | 94.9 | 96.6 | 9449 | 94.6 | 98.4 |
| weather, whether | 61 | 61 | 93.4 | 98.4 | 1226 | 98.4 | 100.0 |
| your, you're | 187 | 103 | 90.4 | 93.6 | 2738 | 90.9 | 99.5 |

Table 1: Pruned versus unpruned performance of Bayes and WinnowS. In the pruned condition, the algorithms use Bayes' usual pruned set of features; in the unpruned condition, they use the full set. The algorithms were trained on 80% of Brown and tested on the other 20%. The 'Features' columns give the number of features used.

| Confusion set | Test cases Within | Test cases Across | Bayes | | WinnowS | |
|---|---|---|---|---|---|---|
| | | | Within | Across | Within | Across |
| accept, except | 50 | 30 | 92.0 | 80.0 | 92.0 | 86.7 |
| affect, effect | 49 | 66 | 98.0 | 84.8 | 100.0 | 93.9 |
| among, between | 186 | 256 | 78.5 | 78.5 | 84.4 | 78.1 |
| amount, number | 123 | 167 | 80.5 | 68.9 | 86.2 | 76.6 |
| begin, being | 146 | 174 | 94.5 | 89.1 | 95.9 | 90.8 |
| cite, sight, site | 34 | 18 | 73.5 | 50.0 | 82.4 | 33.3 |
| country, county | 62 | 71 | 91.9 | 94.4 | 96.8 | 100.0 |
| its, it's | 366 | 1277 | 95.9 | 95.5 | 99.2 | 98.6 |
| lead, led | 49 | 69 | 85.7 | 79.7 | 93.9 | 92.8 |
| fewer, less | 75 | 148 | 92.0 | 94.6 | 93.3 | 93.9 |
| maybe, may be | 96 | 67 | 95.8 | 92.5 | 97.9 | 89.6 |
| I, me | 1225 | 328 | 98.3 | 97.9 | 99.5 | 98.5 |
| passed, past | 74 | 148 | 90.5 | 95.9 | 95.9 | 98.0 |
| peace, piece | 50 | 19 | 92.0 | 78.9 | 94.0 | 89.5 |
| principal, principle | 34 | 30 | 85.3 | 70.0 | 94.1 | 86.7 |
| quiet, quite | 66 | 20 | 89.4 | 60.0 | 90.9 | 75.0 |
| raise, rise | 39 | 118 | 84.6 | 71.2 | 89.7 | 79.7 |
| than, then | 514 | 637 | 93.4 | 96.5 | 97.9 | 98.1 |
| their, there, they're | 850 | 748 | 94.6 | 91.6 | 98.4 | 98.5 |
| weather, whether | 61 | 95 | 98.4 | 94.7 | 100.0 | 96.8 |
| your, you're | 187 | 74 | 90.9 | 85.1 | 99.5 | 97.3 |

Table 2: Across-corpus versus within-corpus performance of Bayes and WinnowS. Training was on 80% of Brown in both cases. Testing for the within-corpus case was on 20% of Brown; for the across-corpus case, it was on 40% of WSJ. The algorithms were run in the unpruned condition.

| Confusion set | Test cases | Bayes | | WinnowS | | |
|---|---|---|---|---|---|---|
| | | Sup only | Sup/unsup | Sup only | Sup/unsup | Incr |
| accept, except | 30 | 80.0 | 86.7 | 86.7 | 93.3 | 86.7 |
| affect, effect | 66 | 84.8 | 90.9 | 93.9 | 92.4 | 92.4 |
| among, between | 256 | 78.5 | 80.5 | 78.1 | 90.6 | 81.2 |
| amount, number | 167 | 68.9 | 77.8 | 76.6 | 88.0 | 79.6 |
| begin, being | 174 | 89.1 | 94.8 | 90.8 | 98.9 | 96.6 |
| cite, sight, site | 18 | 50.0 | 66.7 | 33.3 | 77.8 | 66.7 |
| country, county | 71 | 94.4 | 95.8 | 100.0 | 100.0 | 100.0 |
| its, it's | 1277 | 95.5 | 95.7 | 98.6 | 98.8 | 97.8 |
| lead, led | 69 | 79.7 | 75.4 | 92.8 | 95.7 | 89.9 |
| fewer, less | 148 | 94.6 | 93.2 | 93.9 | 97.3 | 94.6 |
| maybe, may be | 67 | 92.5 | 91.0 | 89.6 | 94.0 | 94.0 |
| I, me | 328 | 97.9 | 98.5 | 98.5 | 98.8 | 98.5 |
| passed, past | 148 | 95.9 | 96.6 | 98.0 | 98.0 | 98.0 |
| peace, piece | 19 | 78.9 | 89.5 | 89.5 | 94.7 | 94.7 |
| principal, principle | 30 | 70.0 | 76.7 | 86.7 | 100.0 | 86.7 |
| quiet, quite | 20 | 60.0 | 70.0 | 75.0 | 80.0 | 80.0 |
| raise, rise | 118 | 71.2 | 87.3 | 79.7 | 90.7 | 84.7 |
| than, then | 637 | 96.5 | 96.2 | 98.1 | 97.6 | 98.3 |
| their, there, they're | 748 | 91.6 | 90.8 | 98.5 | 98.8 | 98.4 |
| weather, whether | 95 | 94.7 | 95.8 | 96.8 | 95.8 | 95.8 |
| your, you're | 74 | 85.1 | 87.8 | 97.3 | 93.2 | 97.3 |

Table 3: Across-corpus performance of Bayes and WinnowS using the sup/unsup strategy, and of WinnowS using incremental learning. Performance is compared with doing supervised learning only. Training in the sup/unsup case is on 80% of Brown plus 60% of WSJ (5% corrupted); in the other cases, it is on 80% of Brown only. Testing in all cases is on 40% of WSJ. The algorithms were run in the unpruned condition.

### 5.3 Dealing with Unfamiliar Test Sets

The preceding section demonstrated that the usual PAC-learning assumption of similar training and test sets may be overly strong for the task of context-sensitive spelling correction. In this section, we present a strategy for dealing with this problem, and evaluate its effectiveness when used by WinnowS and Bayes.

The strategy is based on the observation that the test document, though imperfect, still provides a valuable source of information about its own word usages. Returning to the Desert Storm example, suppose the algorithm is asked to spell-correct an article containing 17 instances of the phrase *Operation Desert Storm*, and 1 instance of the (erroneous) phrase *Operation Dessert Storm*. In this case, the algorithm should be able to learn from the *test* corpus the collocation:

(3)  Operation __ Storm

which tends to imply *desert*. It can then use this feature to fix the one erroneous spelling of the phrase in the test set. It is important to recognize that the system is not "cheating" by looking at the test set; it would be cheating if the system were given an answer key along with the test set.

What the system is really doing is enforcing consistency across the test set. It can detect sporadic errors, but not systematic ones (such as writing *Operation Dessert Storm* every time). However, it should be possible to pick up at least some systematic errors by also doing regular supervised learning on a training set.

This leads to a strategy, which we call *sup/unsup*, of combining supervised learning on the training set with unsupervised learning on the (noisy) test set. We ran both WinnowS and Bayes with the sup/unsup strategy to see the effect on their across-corpus performance. We first needed a test corpus containing errors; we generated one by corrupting a correct corpus. We varied the amount of corruption from 0% to 20%, where a corruption of $p\%$ means we altered a randomly-chosen $p\%$ of the occurrences of the confusion set to be a different word in the confusion set.

The sup/unsup strategy calls for training on both a training corpus and a corrupted test corpus, and testing on the uncorrupted test corpus. For purposes of this experiment, however, we split the test corpus into two parts to avoid any confusion about training and testing on the same data. We trained on 80% of Brown plus a corrupted version of 60% of WSJ; and we tested on the uncorrupted version of the other 40% of WSJ.

The results for the 5% level of corruption are shown in Table 3; this level of corruption corresponds to typical typing error rates.[4] The table compares across-corpus performance of each algorithm with and without the additional boost of unsupervised learning on part of the test corpus. While both Bayes and WinnowS benefit from the unsupervised learning, the effect is particularly strong for WinnowS, which increases its lead over Bayes, outscoring it in the sup/unsup condition (often by a wide margin) on all but one confusion set. This provides impressive evidence of the ability of WinnowS to adapt to the test set using this strategy.

The last column of Table 3 shows an alternative adaptation strategy for WinnowS using incremental learning. In this strategy, WinnowS is trained on 80% of Brown, and tested on 40% of WSJ. After each test instance, WinnowS is allowed to look at the answer and use it as an additional training example. Thus by the end of the test, WinnowS has effectively trained on 80% of Brown plus 40% of WSJ. Not surprisingly, this strategy outperforms training on 80% of Brown only; it is not as good as sup/unsup, although it is not directly comparable, as it trains on different data.

It should be borne in mind that the results in Table 3 depend on two factors. The first is the size of the test set; the larger the test set, the more information it can provide during unsupervised learning. A quantitative analysis of this effect is currently under investigation. The second factor affecting the performance of sup/unsup is the percentage corruption of the test set. Figure 1 shows performance as a function of percentage corruption for a representative confusion set, $\{begin, being\}$. As one would expect, the improvement from unsupervised learning decreases as the percentage corruption increases. For Bayes' performance on $\{begin, being\}$, 20% corruption is almost enough to negate the improvement from unsupervised learning.

## 6 CONCLUSION

While theoretical analyses of the Winnow family of algorithms have predicted an excellent ability to deal with large numbers of features and to adapt to new trends not seen during training, these properties have remained largely undemonstrated. In the work reported here, we have applied a Winnow-based algorithm to context-sensitive spelling correction, a task that addresses an important real-world problem, and that has a potentially huge number of features (over 10,000 in some of our experiments). We compared our Winnow-based algorithm, WinnowS, to a Bayesian statistics-based algorithm representing the state of the art for this task. WinnowS was found to exhibit two striking advantages: first, when the algorithms were run with full feature sets (by doing only minimal pruning of features), WinnowS achieved quite impressive accuracies, outscoring Bayes on 20 out of 21 confusion sets tried. Second, WinnowS was found to be consistently better than Bayes at adapting to an unfamiliar test corpus when using a strategy combining supervised learning on the training set with unsupervised learning on the test set. We believe that these advantages of WinnowS stem from the faster convergence it achieves through its multiplicative weight-update algorithm, and its consequent ability to deal with large numbers of features, including so-called small disjuncts that cover a very small number of instances each.

---
[4]Mays et al. [1991], for example, consider error rates from 0.01% to 10% for the same task.

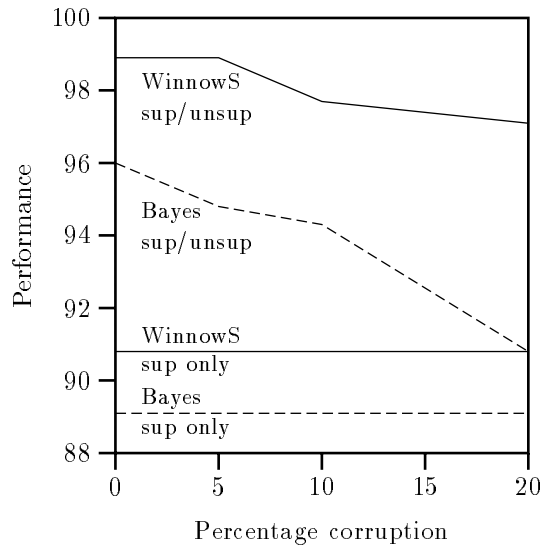

Figure 1: Across-corpus performance of Bayes (dotted lines) and WinnowS (solid lines) with the sup/unsup strategy and with supervised learning only. The curves show performance as a function of the percentage corruption of the test set. Training in the sup/unsup case is on 80% of Brown, plus 60% of WSJ (corrupted); for the supervised-only case, it is on 80% of Brown only. Testing in both cases is on 40% of WSJ. The algorithms were run for the confusion set $\{begin, being\}$ in the unpruned condition.

The Winnow-based approach presented in this paper is being developed as part of a research program in which we are trying to understand how networks of simple and slow neuron-like elements can encode a large body of knowledge and perform a wide range of interesting inferences almost instantaneously. In particular, we investigate this question in a system developed for the purpose of learning knowledge representations for natural language understanding tasks. In light of the encouraging results presented here, we are now extending the approach to other NLP-related tasks.

## Acknowledgements

The second author's research was supported by the Feldman Foundation and a grant from the Israeli Ministry of Science and the Arts; it was done partly while at Harvard University supported by NSF grant CCR-92-00884 and by DARPA AFOSR-F4962-92-J-0466.